\documentclass[11pt]{article}
\usepackage{geometry}
\usepackage{coling2020}
\usepackage{times}
\usepackage{url}
\usepackage{latexsym}
\usepackage{microtype}
\usepackage{floatrow}
\usepackage{graphicx, subfig}
\usepackage{multirow}
\usepackage{array}
\usepackage{xspace}
\usepackage{bbm, amsmath, amssymb}
\usepackage{booktabs}
\usepackage{todonotes}

\usepackage{xcolor}
\usepackage{hyperref}
\definecolor{darkblue}{rgb}{0, 0, 0.5}
\hypersetup{
    colorlinks=true,
    citecolor=darkblue,
    linkcolor=darkblue,
    urlcolor=darkblue,
}

\newcommand{\concat}{; \,}
\newcommand{\boldv}{\mathbf{v}}
\newcommand{\boldu}{\mathbf{u}}
\newcommand{\boldx}{\mathbf{x}}
\newcommand{\boldy}{\mathbf{y}}
\newcommand{\boldh}{\mathbf{h}}
\newcommand{\boldz}{\mathbf{z}}
\newcommand{\bolde}{\mathbf{e}}

\DeclareMathOperator*{\classifier}{Classifier}

\DeclareMathOperator*{\argmax}{arg\;max}

\DeclareMathOperator*{\maxpool}{MaxPool}
\DeclareMathOperator*{\meanpool}{MeanPool}

\newcommand{\context}{\textsc{Context}\xspace}
\newcommand{\original}{\textsc{Original}\xspace}
\newcommand{\ft}{\textsc{Finetune}\xspace}
\newcommand{\fz}{\textsc{Freeze}\xspace}
\newcommand{\extra}{\textsc{Extra}\xspace}
\newcommand{\edit}{\textsc{Edit}\xspace}

\newcommand{\ind}{\mathbbm{1}}

\newfloatcommand{capbtabbox}{table}[][\FBwidth]

\colingfinalcopy %

\title{Duluth at SemEval-2020 Task 7:\\
Using Surprise as a Key to Unlock Humorous Headlines}

\author{
Shuning Jin$^{1}$ \qquad
Yue Yin$^{2}$ \qquad
XianE Tang$^{2}$ \qquad
Ted Pedersen$^{2}$\\
$^{1}$Rutgers University \qquad
$^{2}$University of Minnesota Duluth\\
\small\tt{shuning.jin@rutgers.edu, \{yinxx325,tang0611,tpederse\}@d.umn.edu}
}

\date{}

\begin{document}
\maketitle

\begin{abstract}
We use pretrained transformer-based language models in SemEval-2020 Task 7: Assessing the Funniness of Edited News Headlines. Inspired by the incongruity theory of humor, we use a contrastive approach to capture the surprise in the edited headlines. In the official evaluation, our system gets 0.531 RMSE in Subtask 1, 11\textsuperscript{th} among 49 submissions. In Subtask 2, our system gets 0.632 accuracy, 9\textsuperscript{th} among 32~submissions.
\end{abstract}

\section{Introduction}

\blfootnote{
\hspace{-0.65cm}  %
This work is licensed under a Creative Commons Attribution 4.0 International License. License details: \url{http://creativecommons.org/licenses/by/4.0/}.
}

Humor detection is a challenging problem in natural language processing.
SemEval-2020 Task 7~\cite{SemEval2020Task7}%
\footnote{Competition page:\;\;\url{https://competitions.codalab.org/competitions/20970}}
focuses on detecting humor in English news headlines with \textit{micro-edits}. Specifically, the edited headlines have one selected word or entity that is replaced by editors,
which are
then graded by the degree of funniness. Accurate scoring of the funniness from micro-edits can serve as a footstone of humorous text generation~\cite{SemEval2020Task7}.

Inspired by the incongruity theory~\cite{veale2004incongruity,sep-humor}, we believe that contrast and surprise is a key ingredient of humor. We instantiate this intuition with a contrastive framework. We then systematically compare three widely used models: CBOW, BERT~\cite{devlin-etal-2019-BERT}, and RoBERTa~\cite{liu2019RoBERTa}, providing a benchmark for this task. Our best system, based on RoBERTa, achieves compelling performance for both subtasks.
Our code is available on GitHub.%
\footnote{Code:\;\;\url{https://github.com/dora-tang/SemEval-2020-Task-7}}

\section{Related Work}

Early
humor recognition systems are mostly based on traditional machine learning methods, such as support vector machine, decision tree, Naive Bayes, and $k$-nearest neighbors \cite{castro2016joke}. Besides, an $n$-gram language model shows good performance \cite{yan-pedersen-2017-duluth}
in learning a sense of humor from tweets.
Yet $n$-gram models are limited to a small number of
context words.

Pretrained language models based on Transformer~\cite{vaswani2017attention} can obtain contextual information of a whole sentence.
Among this family, BERT has been used to assess the humor in tweets and jokes
\cite{mao2019bert,weller2019humor}.
Enlightened by these recent advances, we use BERT to
judge the funniness of
edited news headlines. We additionally experiment with RoBERTa, a robustly optimized variant of BERT.

Lastly, several works also attempt to explicitly model incongruity and surprise of humourous text, focusing on homophonic puns.
\newcite{kao2016computational} formalizes incongruity as a mixed effect of ambiguity and distinctiveness, quantified by entropy and Kullback-Leibler divergence. \newcite{he-etal-2019-pun} proposes a local-global surprisal measure based on the log-likelihood ratio, to assess whether a sentence is a pun. However, we focus on a broader definition of humor, and formulate incongruity as an input pair to a dual encoder framework.

\section{Task Data}
The Humicroedit dataset~\cite{hossain-etal-2019-president} provides the training, development, and test data for this task. We also use additional training data from the FunLines dataset~\cite{hossain2020stimulating}.
The dataset statistics are summarized in Appendix~\ref{appendix:dataset}.
In Subtask 1, the goal is predicting the funniness score of an edited headline. The score $z$ ranges from 0 to 3, where 0 means not funny and 3 means very funny. In Subtask 2, the goal is to predict the funnier between an edited sentence pair. For labels $y \in \left\{0,1,2\right\}$, 0 implies two headlines are equally funny,
1 implies the first is the funnier, and 2 implies the second is the funnier.
Examples of the two subtasks are in Table~\ref{tab:sub3}.

\begin{table}[t]%
\centering \fontsize{8.6}{12}\selectfont \setlength{\tabcolsep}{0.5em}
\renewcommand{\arraystretch}{0.88}
\begin{tabular}{llllrr}
\toprule
\textbf{Task} & \textbf{ID} & \textbf{Original Headline} & \textbf{Edit} & \textbf{Score} &  \textbf{Label}  \\
\midrule
\multirow{2}{*}{Subtask 1} 
& 33210 & California and \underline{\textbf{President Trump}} are going to war with each other & monkeys & 1.8 & / \\
\cmidrule{2-6}
& 1664 & What if \underline{\textbf{sociologist}} had as much influence as economists? & donkeys & 2.8 & / \\

\midrule
\multirow{4}{*}{Subtask 2} 
& 9934 & Chibok girls reunited with \underline{\textbf{families}} & smartphones & 1.8 & \multirow{2}{*}{1} \\
& 14279 & Chibok girls reunited with \underline{\textbf{families}} & cats & 0.0 &  \\
\cmidrule{2-6}
& 10920 & Gene Cernan, last \underline{\textbf{astronaut}} on the Moon, dies at 82 & 
dancer & 1.2  &  \multirow{2}{*}{1} \\
& 9866 & Gene Cernan, last astronaut on the Moon, \underline{\textbf{dies}} at 82 & 
impregnated & 0.8 &  \\
\bottomrule
\end{tabular}

\caption{Examples in train data of both subtasks. \underline{Underlined words} in original headlines are to be substituted by the edit words.
}
\label{tab:sub3}
\vspace{-0.2cm}
\end{table}

\section{Methods}
What are the important characteristics of humor? The incongruity theory, a dominant theory of humor, states that ``it is the perception of something incongruous—something that violates our mental patterns and expectations''~\cite{sep-humor}.
Therefore,
we hypothesize an edited headline is funny if the edit words are
semantically
distant from the context words or the original words.
This can be exemplified by the first two examples in Table~\ref{tab:sub3}. We start by looking at Headline 33210 (the \textsc{Monkey Example}), also shown below. The context sentence is extracted from the edit sentence by replacing the edit words with a single \texttt{[MASK]} token, motivated by masked language models. Humans are likely to predict a place or a character for the masked token, while the edit token is \textit{``monkeys''}.

\begin{table}[h]
\vspace{-0.5em}
\centering
\fontsize{9.6}{12}\selectfont \setlength{\tabcolsep}{1em}

\begin{tabular}{lll} %
Original & 
$\underset{1}{\textsc{California}}$ 
$\underset{2}{\textsc{And}}$ 
$\underset{i=3}{\textsc{\underline{President}}}$ 
$\underset{j=4}{\textsc{\underline{Trump}}}$ 
$\underset{5}{\textsc{Are}}$ 
$\underset{6}{\textsc{Going}}$ 
$\underset{7}{\textsc{To}}$ 
$\underset{8}{\textsc{War}}$ 
& $ T_o =8$ 
\\
Edit &
$\underset{1}{\textsc{California}}$ 
$\underset{2}{\textsc{And}}$ 
$\underset{i=3,\;k=3}{\textsc{\underline{Monkeys}}}$ 
$\underset{4}{\textsc{Are}}$ 
$\underset{5}{\textsc{Going}}$ 
$\underset{6}{\textsc{To}}$ 
$\underset{7}{\textsc{War}}$ 
& $  T_e =7$
\\
Context  &
$\underset{1}{\textsc{California}}$ 
$\underset{2}{\textsc{And}}$ 
$\underset{i=3}{\textsc{\underline{[Mask]}}}$ 
$\underset{4}{\textsc{Are}}$ 
$\underset{5}{\textsc{Going}}$ 
$\underset{6}{\textsc{To}}$ 
$\underset{7}{\textsc{War}}$ 
& $ T_c =7$
\\
\end{tabular}

\vspace{-0.5em}
\label{tab:example_motivating}
\end{table}

Similarly for Headline 1664, given the context \textsc{What If [MASK] Had As Much Influence As Economists}, humans might fill in occupation-related words like \textit{``scientist''} or \textit{``sociologist''} (as in the original headline). However, the edit word is \textit{``donkeys''}, which is a surprising prediction and is considered very funny (scored 2.8 out of 3). In this section, we describe a concrete architecture that models the strength of contrast and surprise, which translates into the funniness score.

\subsection{Span Representation}\label{sec:span}

\begin{figure*}[t]
\centering
\includegraphics[scale=0.23]{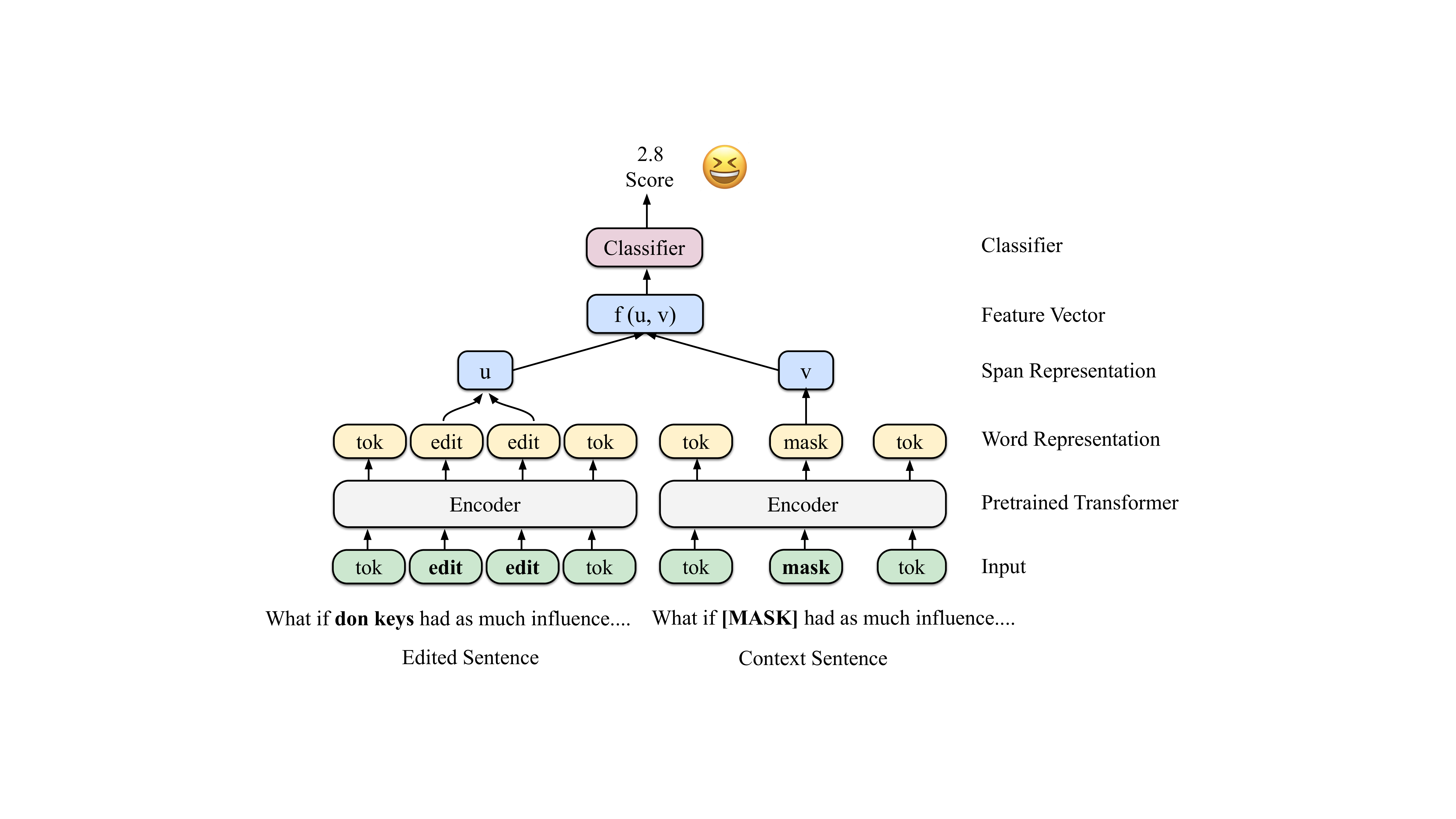}
\caption{Transformer architecture to predict the funniness score of an edited headline using edit-context sentence pair. The full edited sentence is \textit{``What if donkeys had as much influence as economists''}. \textit{``donkeys''} is the edited word and is tokenized into two subwords in this example.
}
\label{fig:architecture}
\end{figure*}

Let
$\boldx = (x_1 \ldots x_{T_o})$,
$\Tilde{\boldx} = (\Tilde{x}_1 \ldots \Tilde{x}_{T_e})$,
$\boldx^\star = (x^\star_1 \ldots x^\star_{T_c})$
denote the original, edit, context token sequences.
A pretrained word embedding or pretrained encoder maps the tokens into vector sequences $\bolde_{1:T_o}$, $\Tilde{\bolde}_{1:T_e}$, $\bolde^\star_{1:T_c}$.
The goal is to encode edit sentence, original sentence, context sentence into fixed-length vector representations $\boldu$, $\boldv'$, $\boldv$  $\in \mathbb{R}^{d}$. Importantly, we use span (a.k.a. sub-sentence) representation rather than whole sentences, which corresponds to the underlined ranges in the above \textsc{Monkey Example}.
Denote a span as a tuple of start and end position of contiguous tokens: $s = [i, j]$ for edit, $ \Tilde{s}  = [i, k]$ for original, and $ s^\star= [i, i]$ for the \texttt{[MASK]} token.

\paragraph{CBOW}
We first explore context-independent word representations. We use pretrained GloVe~\cite{pennington-etal-2014-glove} vectors with $d=300$ and a vocabulary of 2.2 million words. We use word averaging to get edit word vector
$\boldu = \meanpool(\Tilde{\bolde}_{i} \ldots \Tilde{\bolde}_{k})$ and original word vector $\boldv' = \meanpool(\bolde_i \ldots  \bolde_j) $. We max pool all context words to extract the most salient features. The context vector is
$\boldv = \maxpool(\bolde^\star_{1}\ldots\bolde^\star_{i-1}, \bolde^\star_{i+1}\ldots\bolde^\star_{T_c})$.

\paragraph{Transformer}
We use pretrained transformer-based language models to obtain contextual word representations. The architecture is shown in Figure~\ref{fig:architecture}, a Siamese network \cite{bromley1994signature} where the two encoders have identical structures and shared parameters.%
\footnote{An alternative architecture is to use a single transformer to encode the concatenation of a sentence pair, which enforces cross-sentence attention. However, since the paired sentences here are almost identical, cross-sentence attention seems unnecessary.}
With the self-attention mechanism, each word attends to all other words in the sentence and aggregates contextual information.
The edit and original vectors are obtained by averaging:
$\boldu = \meanpool(\Tilde{\bolde}_{i} \ldots \Tilde{\bolde}_{k})$,
$\boldv' = \meanpool(\bolde_i \ldots \bolde_j)$.
The context vector is simply the contextual embedding of the masked token: $\boldv = \bolde^\star_i$.
We experiment with BERT and RoBERTa, using the PyTorch~\cite{paszke2019pytorch} implementation from HuggingFace Transformers library~\cite{Wolf2019HuggingFacesTS}.%
\footnote{HuggingFace Transformers library:\;\;\url{https://github.com/huggingface/transformers}}
We use \texttt{bert-base-uncased} ($L=12$, $d=768$, lower-cased) and \texttt{roberta-base} ($L=12$, $d=768$).

\paragraph{Transfer Paradigm}
When using those pretrained word representations, we consider two transfer paradigms: finetuning (\ft) and
not finetuning (\fz).
In the case of \fz, we use fixed word embedding directly as the feature for CBOW. For transformers, we use a weighted average of hidden layers %
from the frozen encoder, with trainable mixing scalars. This approach follows ELMo \cite{peters-etal-2018-deep} and the edge probing model \cite{tenney2019learn}. Specifically, the final aggregated embedding for $i^{\text{th}}$ position is
$\bolde_i = \gamma \sum_{l=0}^L \alpha_l \; \bolde_{l,i} $,
where $\gamma$ is a scaling factor, $\alpha_l$ is the weight of the $l^{\text{th}}$ layer, $L$ is the total number of layers, and $l=0$ corresponds to the embedding layer.

\subsection{Task Specific}

\paragraph{Regression}
As mentioned at the beginning of the section, contrast and surprise is the key to humor. To represent the pairwise relationship between two vectors, we derive feature from
$ \boldh = f(\boldx, \boldy) = [\boldx \concat \boldy \concat \left|{\boldx - \boldy}\right| \concat \boldx * \boldy ] \in \mathbb{R}^{4d}$,
where $\concat$ denotes concatenation and $*$ denotes element-wise multiplication. This feature has been used as the input to the classifier in the sentence pair tasks of SentEval \cite{conneau-kiela-2018-senteval}. To formulate the contrast pair, we either use edit sentence and its context $f(\boldu, \boldv)$, or edit sentence and original sentence $f(\boldu,\boldv')$. We denote the two scenarios as \context and \original respectively.
Finally, we use a classifier to predict the funniness score of the edited headline.
$\hat{z}=\classifier(\boldh) \in \mathbb{R}$.
The classifier is a two-layer MLP with 256 hidden dimensions.
When finetuning transformers we use single-layer linear projection instead,
since its large number of parameters have already given us sufficient flexibility.
The optimization objective is mean squared error
$\mathcal{L} = \Vert\boldz-\hat{\boldz}\Vert^{2}$.

\paragraph{Classification}
In Subtask 2, we use the same method to predict the scores of two edited versions $\hat{z}^{(1)}$ and $\hat{z}^{(2)}$. By comparing the scores, the funnier version is found during evaluation and testing time:
$\hat{y} = \argmax_{i\in \{1,2\}} \hat{z}^{(i)}$.
The loss function is
$\mathcal{L} = \Vert\boldz^{(1)}-\hat{\boldz}^{(1)}\Vert^{2} + \Vert\boldz^{(2)}-\hat{\boldz}^{(2)}\Vert^{2}$.

\begin{table}[t]
\centering \fontsize{9.0}{12}\selectfont \setlength{\tabcolsep}{0.5em}
\renewcommand{\arraystretch}{0.88}
\begin{tabular}{@{}l|crc|crrc@{}}
\toprule
{} & \multicolumn{3}{c}{\textbf{Subtask 1}} & \multicolumn{4}{c}{\textbf{Subtask 2}} \\ \textbf{Model} & \textbf{RMSE$^\dagger$} & \textbf{Gain} & \textbf{Spearman} & \textbf{Accuracy$^\dagger$} & \textbf{Gain} & \textbf{Reward} & \textbf{RMSE} \\
\midrule
Baseline & 0.575 & -0.033 & / & 0.490 & -0.109 & -0.020 & / \\
\midrule
CBOW &&&&&&& \\
\quad with \textsc{Context}+\textsc{Freeze} & \textbf{0.542} & \textbf{0.000} & \textbf{0.319} & 0.599 & 0.000 & 0.184 & 0.546 \\
\quad +\textsc{Original} & 0.559 & -0.017 & 0.243 & 0.599 & 0.000 & 0.169 & 0.760 \\
\quad +\textsc{Extra} & 0.544 & -0.002 & 0.308 & 0.605 & 0.006 & \textbf{0.191} & \textbf{0.545} \\
\quad +\textsc{Original}+\textsc{Extra} & 0.558 & -0.016 & 0.250 & 0.601 & 0.002 & 0.173 & 0.574 \\
\quad +FT & 0.544 & -0.002 & 0.311 & 0.604 & 0.005 & 0.178 & \textbf{0.545} \\
\quad +FT+\textsc{Original} & 0.561 & -0.019 & 0.247 & 0.592 & -0.007 & 0.165 & 0.604 \\
\quad +FT+\textsc{Extra} & 0.548 & -0.006 & 0.313 & \textbf{0.606} & \textbf{0.007} & 0.188 & 0.547 \\
\quad +FT+\textsc{Original}+\textsc{Extra} & 0.563 & -0.021 & 0.261 & 0.589 & -0.010 & 0.161 & 0.581 \\
\midrule
BERT &&&&&&& \\
\quad with \textsc{Context}+\textsc{Freeze} & 0.531 & 0.011 & 0.384 & 0.616 & 0.017 & 0.207 & 0.546 \\
\quad +\textsc{Original} & 0.534 & 0.008 & 0.366 & 0.603 & 0.004 & 0.186 & 0.547 \\
\quad +\textsc{Extra} & \textbf{0.530} & \textbf{0.012} & 0.382 & 0.615 & 0.016 & 0.207 & \textbf{0.530} \\
\quad +\textsc{Original}+\textsc{Extra} & 0.541 & 0.001 & 0.346 & 0.615 & 0.016 & 0.204 & 0.542 \\
\quad +FT & 0.536 & 0.006 & 0.366 & \textbf{0.635} & \textbf{0.036} & 0.234 & 0.552 \\
\quad +FT+\textsc{Original} & 0.536 & 0.006 & 0.365 & 0.628 & 0.029 & 0.231 & 0.555 \\
\quad +FT+\textsc{Extra} & 0.541 & 0.001 & 0.373 & 0.630 & 0.031 & 0.232 & 0.554 \\
\quad +FT+\textsc{Original}+\textsc{Extra} & 0.533 & 0.009 & \textbf{0.387} & 0.629 & 0.030 & \textbf{0.236} & 0.550 \\
\midrule
RoBERTa &&&&&&& \\
\quad with \textsc{Context}+\textsc{Freeze} & 0.528 & 0.014 & 0.388 & 0.635 & 0.036 & 0.246 & \underline{\textbf{0.529}} \\
\quad +\textsc{Original} & 0.536 & 0.006 & 0.366 & 0.625 & 0.026 & 0.224 & 0.540 \\
\quad +\textsc{Extra} & 0.528 & 0.014 & 0.390 & 0.640 & 0.041 & 0.252 & 0.536 \\
\quad +\textsc{Original}+\textsc{Extra} & 0.533 & 0.009 & 0.368 & 0.618 & 0.019 & 0.207 & 0.543 \\
\quad +FT & 0.534 & 0.008 & 0.382 & 0.649 & 0.050 & \underline{\textbf{0.254}} & 0.535 \\
\quad +FT+\textsc{Original} & 0.527 & 0.015 & \underline{\textbf{0.425}} & \underline{\textbf{0.650}} & \underline{\textbf{0.051}} & \underline{\textbf{0.254}} & 0.538 \\
\quad +FT+\textsc{Extra} & 0.526 & 0.016 & 0.407 & 0.638 & 0.039 & 0.233 & 0.594 \\
\quad +FT+\textsc{Original}+\textsc{Extra} & \underline{\textbf{0.522}} & \underline{\textbf{0.020}} & 0.410 & 0.626 & 0.027 & 0.216 & 0.625 \\
\bottomrule
\end{tabular}
\caption{Post-evaluation test results.
\fz: no finetuning; FT: finetuning; \context: use context of edit headline; \original: use original headline; \extra: use additional training data from the FunLines dataset.
\textbf{RMSE}$^\dagger$ and \textbf{Accuracy}$^\dagger$ are primary metrics. The best within each model type is \textbf{bolded}, and the overall best is \underline{underlined}.
Gain measures the improvement over CBOW-\context-\fz. In Subtask 1 this is w.r.t. to RMSE and in Subtask 2 w.r.t. accuracy.
Baseline 1 uses the average score and Baseline 2 uses the majority label.
}
\label{tab:result}
\end{table}

\section{Experiments and Results}

\subsection{Metrics}
For Subtask 1, the primary metric for official ranking is Root Mean Squared Error (RMSE). In addition, we calculate Spearman's rank correlation coefficient which measures the monotonic relationship between predicted scores and true scores.
In the evaluation of Subtask 2, instances with label 0 are ignored. The primary metric for official ranking is accuracy.
As an auxiliary metric, reward takes pairwise score differences into account:
$r = \frac{1}{N} \sum_{i=1}^N (\ind_{\hat{y}_i=y_i} - \ind_{\hat{y}_i \ne y_i}) |z_i^{(1)}-z_i^{(2)}|$, where $y_i$ and $\hat{y}_i$ are true labels and predicted labels respectively, and $z_i^{(1)}$ and $z_i^{(2)}$ are true scores.

\subsection{Official Evaluation}
For the official evaluation, our submitted system is RoBERTa-\fz-\context.
We use Adam \cite{kingma2014adam} optimizer with a learning rate of 1e-3 and use the 10\textsuperscript{th} epoch. Our system gets 0.531 RMSE for Subtask 1 (11$^\text{th}$ among 49 submissions) and 0.632 accuracy for Subtask 2 (9$^\text{th}$ among 32 submissions) on the test set.%
\footnote{Task leaderboard:\;\;\url{https://competitions.codalab.org/competitions/20970\#results}. ``Evaluation-Task-1" is for Subtask 1 and ``Evaluation-Task-2" is for Subtask 2.}

\subsection{Post-Evaluation}
In the post-evaluation phase, we conduct a more extensive search on hyperparameters and select the best models based on validation performance. Experiment details and hyperparameters are in Appendix~\ref{appendix:details}.
We systematically compare CBOW, BERT, and RoBERTa, and perform an ablation study to understand the effects of various factors: extra training data, finetuning or freezing the pretrained embeddings, and using \context or \original feature.
The post-evaluation results on the test set are in Table~\ref{tab:result}.

\paragraph{Contextual Representation}
Despite its simplicity, CBOW is surprisingly effective. Its best result is significantly better than the baseline and is comparable to Subtask 1 \#19 (0.547) and Subtask 2 \#17 (0.605) on the leaderboard. By comparing the three models, we see that pretrained language models have better performance than context-independent word embedding. While results for BERT and RoBERTa are similar, both of them outperform CBOW, evidencing that contextual information is essential for humor detection.

\paragraph{\context vs. \original}
In the ablation study, we first notice that neither finetuning nor using extra data from the FunLines dataset make much difference for all models.
Interestingly, using different
contrast pairs
as the feature has different effects on models. \context is better than \original for CBOW, yet they are similar for pretrained language models.
Why does this happen?
CBOW with \original only uses the information of edited word and original word while completely neglecting the contextual relation. Pairing CBOW with \context can alleviate this limitation. On the other hand, pretrained language models exploit the contextual relation between edit words and context words in both cases.

\section{Analysis and Discussion}

\subsection{Non-contrastive Approach}
In our main experiment, we focus on the contrastive approach using a sentence pair (i.e., \context and \original) and show its effectiveness. The remaining question is, can we predict humor using
the edited sentence
as the only input?
Thus, we investigate a non-contrastive approach, with a single encoder to obtain the span representation of the edit sentence. We refer to this as \edit. This is equivalent to only using the left part in Figure~\ref{fig:architecture}. We conduct an experiment with RoBERTa on both subtasks. The results are in Table~\ref{tab:feature}. We see that \edit has similar performance as contrastive approaches. We conjecture that \edit captures contrast \textit{implicitly}, while \context and \original capture contrast \textit{explicitly} by design.

\begin{table}[t]
\centering \fontsize{8.6}{12}\selectfont \setlength{\tabcolsep}{0.5em}

\begin{tabular}{lcc}
\toprule
\textbf{Feature} &  \textbf{Subtask 1 RMSE} &  \textbf{ Subtask 2 Accuracy} \\
\midrule
\context  & 0.526  &  0.649 \\
\original  &  0.522 &  0.650 \\
\midrule
\edit  &  0.526  &  0.651 \\
\bottomrule
\end{tabular}

\caption{
Comparison between non-contrastive and contrastive approaches, based on RoBERTa.
\context and \original are contrastive, using the edit-context pair and the edit-original pair as input respectively. \edit is non-contrastive, using the edit sentence only.  The results are from test sets of the two subtasks. We tune other hyperparameters on the validation sets and select the best model for each cell.
}
\label{tab:feature}
\end{table}

\begin{table}[t]%
\centering\small
\definecolor{Orange}{cmyk}{0,0.61,0.87,0}
\definecolor{ForestGreen}{cmyk}{0.91,0,0.88,0.12}

\newcommand{\colorpearson}{\textcolor{ForestGreen}}
\newcommand{\colorspearman}{\textcolor{Orange}}

\begin{tabular}{@{}lcccc@{}}
\toprule
& \textbf{Human} &  \textbf{\original} &  \textbf{\context} & \textbf{\edit} \\

\midrule
Human  & /   & \colorspearman{0.41 }  &  \colorspearman{0.41}  &  \colorspearman{0.42} \\
\original  &  \colorpearson{0.42}  & /  &  \colorspearman{0.84}  &    \colorspearman{0.77} \\
\context  &   \colorpearson{0.41}  &  \colorpearson{0.84} & /  &  \colorspearman{0.80} \\
\edit      &  \colorpearson{0.43}  & \colorpearson{0.79}  &  \colorpearson{0.81} & /   \\

\bottomrule
\end{tabular}

\caption{The correlation matrix of human scores and predicted scores from RoBERTa with different features: \edit, \context, and \original.
Lower triangle: \colorpearson{Pearson} correlation coefficients. Upper triangle: \colorspearman{Spearman} correlation coefficients.
Results are from the test set of Subtask 1.}
\label{tab:correlation_both}
\end{table}

\subsection{Error Analysis}

To understand the relationship between human judgment and model predictions, we calculate the correlation matrix between true funniness scores and predicted scores from RoBERTa with different features (\context, \original, and \edit) for Subtask 1. From Table~\ref{tab:correlation_both}, we see that the models correlate poorly with human judgment (correlations $\approx$ 0.4), while correlating well with each other (correlation $\approx$ 0.8).

To further learn when the models make an erroneous judgment, we look at model predictions on the test set of Subtask 1. We see the models generally capture the incongruity phenomenon.
While being key to many examples, incongruity does not account for others.
We summarize some typical examples in Table~\ref{tab:human_vs_model}:
\begin{itemize}
\item
For Headline 9100, the edit word \textit{``children''} is incongruous with the context of national security. While humans consider it not funny at all, models assign a high funniness score. The fallacy is that incongruity is not a sufficient condition for humor.
\item
In other cases, the edit words are congruous with the context. While humans consider them very funny, models predict the opposite. That is, incongruity is not a necessary condition for humor.
Humor has diverse underlying causes. For instance, Headline 12685 shows \textit{sarcasm}, taunting Trump's lack of geography knowledge and common sense. Headline 12271 uses \textit{pun} based on polysemy: \textit{``turkey''} can either mean a country (when capitalized) or a bird.
Also, humor can require an understanding of \textit{cultural commentary}, exemplified by Headline 9406. Since the Cheesecake Factory is a large chain of restaurants that some may look down upon, they are happy to see it blown up with \textit{``no complaints''}.
\end{itemize}

\begin{table}[t]%
\centering \fontsize{8.6}{12}\selectfont \setlength{\tabcolsep}{0.5em}
\newcolumntype{C}[1]{>{\centering\arraybackslash}m{#1}}
\newcolumntype{L}[1]{>{\raggedright\arraybackslash}m{#1}}
\newcolumntype{R}[1]{>{\raggedleft\arraybackslash}m{#1}}
\renewcommand{\arraystretch}{0.88}

\begin{tabular}{@{}L{1cm}L{6.4cm}L{1cm}C{1.2cm}C{1.2cm}C{1.2cm}C{1cm}@{}}
\toprule
\textbf{ID} & \textbf{Original Headline} & \textbf{Edit} & \textbf{Human} & \multicolumn{3}{c}{\textbf{Prediction}} \\
& & & &  \original & \context & \edit \\

\midrule
9100 &
WSJ: Trump's top national security adviser is being investigated for his communications with \underline{\textbf{Russia}}
& children & 0.0 & 1.35 & 1.50 & 0.94 \\

\midrule
9406 &
Man Sets Off Explosive Device at L.A.-Area Cheesecake Factory, No \underline{\textbf{Injuries}}
& complaints & 2.4 & 0.76 & 0.52 & 0.72 \\

\midrule
12271 &
Turkey tells \underline{\textbf{citizens}} to reconsider travelling to US
& poultry & 2.4 & 0.85 & 0.83 & 0.65 \\

\midrule
12685 &
CBS Poll: Americans lack confidence in Trump's ability to \underline{\textbf{handle}} North Korea
& locate & 2.4 & 0.97 & 0.94 & 0.96 \\

\bottomrule
\end{tabular}

\caption{Error analysis on the test set of Subtask 1. The examples show disagreements between human ratings and model predictions.
}
\label{tab:human_vs_model}
\vspace{-0.2cm}
\end{table}

\section{Conclusions}

We use incongruity as the key to assessing funniness in edited news headlines. Specifically, we use pretrained transformer-based language models to encode contrastive pairs. Our best performing model is RoBERTa, which is submitted for the official evaluation and achieves competitive performance in both subtasks. The additional experiment shows that a non-contrastive approach may also encode incongruity implicitly.
While incongruity is a common ingredient of humor, error analysis indicates it is neither sufficient nor necessary. This invites future research to take other factors (e.g., sarcasm, pun, or world knowledge) into account to better tackle humor, an intricate phenomenon rooted in human creativity.

\section*{Acknowledgements}
The authors would like to thank Karl Stratos for his insightful feedback.

\bibliographystyle{coling}
\bibliography{semeval2020}

\newpage

\appendix

\section{Task Summary}
\label{appendix:dataset}

\begin{table}[h]
\centering \fontsize{9.6}{12}\selectfont \setlength{\tabcolsep}{0.5em}
\begin{tabular}{lllcccc}
\toprule
\textbf{Task} & \textbf{Type} & \textbf{Metric} & \textbf{Train} & \textbf{Train Extra} & \textbf{Dev} & \textbf{Test}\\
\midrule
Subtask 1 & Regression & RMSE & 9653 & 8248  & 2420  & 3025 \\
Subtask 2 & Classification & Accuracy & 9382 & 1959 & 2356 & 2961\\
\bottomrule
\end{tabular}

\caption{Summary of subtasks. Train, Dev, Test are from the Humicroedit dataset. Train Extra is from the FunLines dataset.}
\label{tab:datasets}
\end{table}

\section{Experiment Details}
\label{appendix:details}

\paragraph{Preprocessing}
We use spaCy word tokenizer for CBOW. The pretrained transformers use byte-pair encoding \cite[BPE]{sennrich-etal-2016-neural} to convert text into subword units. BERT uses WordPiece \cite{wu2016googles} tokenization, a character-level BPE, with a vocabulary size of 30K. RoBERTa preserves cases and uses a byte-level BPE with a vocabulary size of 50K.

\paragraph{Training}
For training, we use a batch size of 32 in Subtask 1 and 16 in Subtask 2. We use Adam optimizer and perform gradient clipping with a max $\mathcal{L}_2$ norm of 5.
For most experiments, we train for 10 epochs with a learning rate in \{1e-3, 3e-4\}. However, when finetuning transformers, we choose max epochs in \{3, 10\}, and use either a constant learning rate or a linear decreasing schedule with an initial learning rate in \{2e-5, 5e-5\}. We perform validation on the development set every 1/3 epoch and save the best checkpoint.

\end{document}